\documentclass[letterpaper, 10 pt, conference]{ieeeconf}  

\IEEEoverridecommandlockouts    

\usepackage{graphics} 
\usepackage{epsfig} 
\usepackage{mathptmx} 
\usepackage{times} 
\usepackage{amsmath} 

\usepackage{amssymb}  
\usepackage{cite}
\usepackage[colorlinks,linkcolor=red,anchorcolor=blue,citecolor=green]{hyperref}
\usepackage[capitalize]{cleveref}
\usepackage{multirow}
\usepackage{balance}

\title{\LARGE \bf
HGSLoc: 3DGS-based Heuristic Camera Pose Refinement
}

\author{Zhongyan Niu$^{1}$, Zhen Tan$^{1}$, Jinpu Zhang$^{1}$, Xueliang Yang$^{1}$, Dewen Hu$^{*1}$
  \thanks{$^{1}$the College of Intelligence Science and Technology, National University of Defense Technology, China.}
  \thanks{* indicates corresponding authors: D. Hu (dwhu@nudt.edu.cn)}
  \thanks{This work has partially been funded by the Science and Technology Innovation Program of Hunan Province (2024QK2006).}
}

\begin{document}

\crefname{section}{Sec.}{Secs.}
\Crefname{section}{Section}{Sections}
\Crefname{table}{Table}{Tables}
\crefname{table}{Tab.}{Tabs.}

\maketitle
\thispagestyle{empty}
\pagestyle{empty}

\begin{abstract}

Visual localization refers to the process of determining camera poses and orientation within a known scene representation.
This task is often complicated by factors such as changes in illumination and variations in viewing angles. 
In this paper, we propose HGSLoc, a novel lightweight plug-and-play pose optimization framework, which integrates 3D reconstruction with a heuristic refinement strategy to achieve higher pose estimation accuracy.
Specifically, we introduce an explicit geometric map for 3D representation and high-fidelity rendering, allowing the generation of high-quality synthesized views to support accurate visual localization. Our method demonstrates higher localization accuracy compared to NeRF-based neural rendering localization approaches. We introduce a heuristic refinement strategy, its efficient optimization capability can quickly locate the target node, while we set the step-level optimization step to enhance the pose accuracy in the scenarios with small errors. With carefully designed heuristic functions, it offers efficient optimization capabilities, enabling rapid error reduction in rough localization estimations. Our method mitigates the dependence on complex neural network models while demonstrating improved robustness against noise and higher localization accuracy in challenging environments, as compared to neural network joint optimization strategies. The optimization framework proposed in this paper introduces novel approaches to visual localization by integrating the advantages of 3D reconstruction and the heuristic refinement strategy, which demonstrates strong performance across multiple benchmark datasets, including 7Scenes and Deep Blending dataset. The implementation of our method has been released at \url{https://github.com/anchang699/HGSLoc}.

\end{abstract}

\section{INTRODUCTION}

Visual localization is a research direction that aims to determine the pose and orientation of a camera within a known scene by analyzing and processing image data. This technique has significant applications in various fields, such as augmented reality (AR), robotic navigation, and autonomous driving. By enabling devices to accurately identify their spatial location in complex 3D environments, visual localization facilitates autonomous navigation, environmental awareness, and real-time interaction. The core objective of visual localization is to estimate the camera's absolute pose. However, this task is challenging due to factors such as illumination changes, dynamic occlusions, and variations in viewing angles, necessitating the development of robust and efficient algorithms to address these complexities.

Two major categories of visual location methods are Absolute Pose Regression (APR)\cite{kendall2015posenet,kendall2016modelling,kendall2017geometric,wang2020atloc,chen2021direct,chen2022dfnet,shavit2021learning,chen2024map} and Scene Coordinate Regression (SCR)\cite{brachmann2021visual,brachmann2023accelerated,wang2024glace}. APR is an end-to-end deep learning approach that directly regresses the camera's pose from the input image. The key advantages of APR lie in its simplicity and computational efficiency. However, APR exhibits notable limitations, particularly in complex or previously unseen environments, where its generalization capability is weak\cite{sattler2019understanding}. In contrast, SCR adopts an indirect strategy for pose estimation. It first predicts the 3D scene coordinates of each image pixel using a deep learning model, followed by the computation of the camera pose through the spatial transformation of these coordinates. Although SCR demonstrates high accuracy and robustness in familiar scenes, it incurs substantial computational costs due to the need to predict a large number of pixel-wise coordinates.

\begin{figure}[t]
  \centering
  \includegraphics[width=\linewidth]{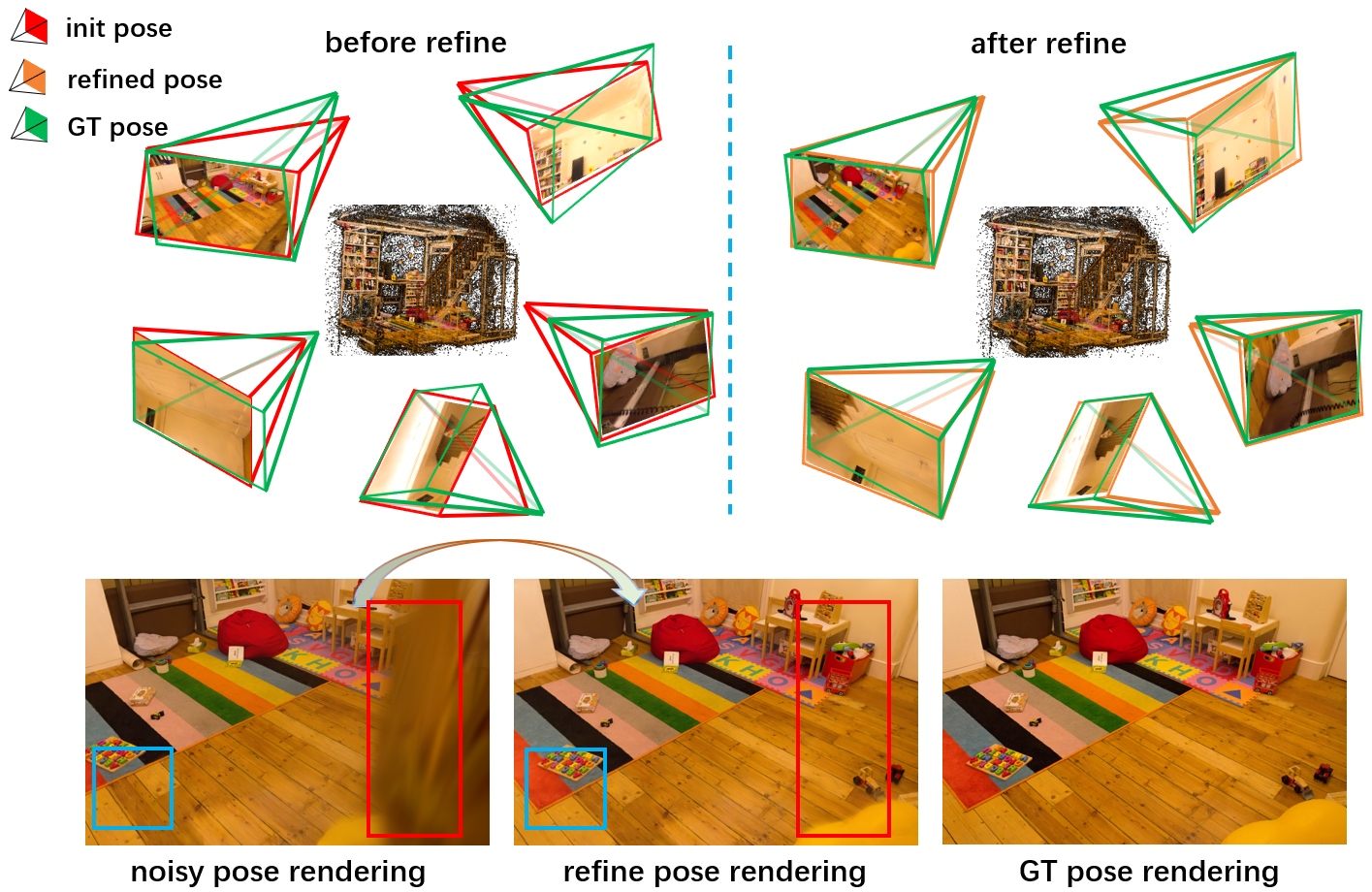}
  \caption{HGSLoc significantly reduces the error between the coarse pose and the GT, and exhibits strong noise resistance.}
  \label{fig:graph1}
  \vspace{-0.5cm}
\end{figure}

In this paper, we propose a novel visual localization paradigm that enhances pose estimation accuracy through the integration of 3D reconstruction. Neural Radiance Field (NeRF)\cite{mildenhall2021nerf} is capable of synthesizing and rendering high-quality 3D scene images through neural network training. Some existing NeRF-based visual localization methods\cite{yen2021inerf,maggio2023loc} have gained widespread recognition. However, NeRF’s pixel-wise training and inference mechanism results in significant computational costs, limiting its practical applications. In contrast, 3D Gaussian Splatting (3DGS)\cite{kerbl20233d} represents scene points as Gaussian distributions, significantly reducing computational overhead and leveraging CUDA acceleration for efficient training and inference. In known or partially known static environments, several approaches, such as 3DGS-ReLoc\cite{jiang20243dgs} and GSLoc\cite{liu2024gsloc}, have been developed. The 3DGS-ReLoc method relies on grid search with the normalized cross-correlation (NCC)\cite{hiasa2018cross}, which affects the localization accuracy. The GSLoc method requires several rounds of iterations to achieve the desired result in the case of poor initial pose estimation, and each iteration utilizes MASt3R\cite{leroy2024grounding} for assisted localization. Whereas, our method is a lightweight framework that enables efficient positional optimization for any image. As shown in \cref{fig:graph1}, by incorporating 3DGS, richer geometric information is available for pose estimation, and through heuristic optimization of coarse pose estimates, the accuracy of localization can be significantly enhanced in complex scenes.

Absolute Pose Regression (APR) and Scene Coordinate Regression (SCR) provide coarse pose estimates for further refinement. To enhance scene rendering, we introduce 3D Gaussian Splatting (3DGS), which constructs a dense point cloud for high-fidelity reconstruction. Building on this, we employ a heuristic refinement algorithm\cite{bulitko2007graph} that efficiently adjusts the rendered view to better align with the query image, improving pose accuracy. This modular approach reduces reliance on computationally expensive neural network training, offering a more efficient alternative to deep learning-based pose optimization. Our method demonstrates strong generalization capabilities, maintaining high accuracy even in the presence of noisy pose data, making it adaptable across various platforms. Experimental results on benchmark datasets, including 7Scenes and Deep Blending, validate its effectiveness in visual localization tasks. The contributions of our approach are summarized as follows:

\begin{itemize}

\item We propose a lightweight, plug-and-play pose optimization framework that facilitates efficient pose refinement for any query image.
\item We design a heuristic refinement strategy and set the step-level optimization step to adapt various complex scenes.
\item Our proposed framework achieves higher localization accuracy than NeRF-based neural rendering localization approach~\cite{zhou2024nerfect} and outperforms neural network joint pose optimization strategy in noisy conditions.

\end{itemize}

\section{RELATED WORK}

In this section, we introduce visual localization methods and 3D Gaussian Splatting.

\vspace{-0.1cm}
\subsection{Visual localization}
PoseNet pioneered Absolute Pose Regression (APR)\cite{kendall2015posenet,kendall2016modelling,kendall2017geometric,wang2020atloc,chen2021direct,chen2022dfnet,shavit2021learning,chen2024map} by employing a convolutional neural network (CNN) to directly regress camera pose from images, bypassing traditional feature extraction and geometric computations. This end-to-end approach simplifies visual localization across diverse environments. MS-Transformer\cite{shavit2021learning} improves APR by incorporating global context modeling and multi-head self-attention, enhancing scene understanding and pose accuracy. DFNet\cite{chen2022dfnet} further extends APR by integrating multimodal sensor data, increasing robustness. However, APR methods remain sensitive to noise and environmental variability, with accuracy degrading under poor lighting, adverse weather, or occlusions.

Scene Coordinate Regression (SCR)\cite{brachmann2021visual,brachmann2023accelerated,wang2024glace} estimates camera pose by mapping image pixels to 3D scene coordinates, eliminating the need for complex feature matching and improving efficiency. DSAC*\cite{brachmann2021visual} enhances SCR with a differentiable hypothesis selection mechanism and supports both RGB and RGB-D inputs, leveraging depth information for improved scene interpretation. ACE\cite{brachmann2023accelerated} optimizes image coordinate encoding and decoding to accelerate feature matching while enhancing robustness to noise and lighting variations, ensuring reliable pose estimation in challenging environments.

\subsection{3D Gaussian Splatting}
 3D Gaussian Splatting (3DGS)\cite{kerbl20233d}, an emerging method in 3D reconstruction, has rapidly gained prominence since its introduction. This method significantly accelerates the synthesis of new views by modeling the scene with Gaussian ellipsoids and utilizing advanced rendering methods. Within the realm of 3DGS research, various techniques have enhanced and optimized 3DGS in different aspects, such as quality improvement\cite{song2024sa}, compression and regularization\cite{lee2024compact}, navigation\cite{chen2024splat}, dynamic 3D reconstruction\cite{wu20244d}, and handling challenging inputs\cite{zhu2023fsgs}. The advancement of 3DGS methods not only enhances the quality of scene reconstruction but also speeds up rendering, offering novel and improved approaches for visual localization tasks. For instance, GSLoc\cite{liu2024gsloc} leverages rendered images from new viewpoints for matching and pose optimization, while the InstantSplat\cite{fan2024instantsplat} method, utilizing DUSt3R\cite{wang2024dust3r}, achieves rapid and high-quality scene reconstruction by jointly optimizing poses with 3D Gaussian parameters. Our proposed method builds upon 3DGS reconstructed scenes and employs heuristic pose optimization to enhance pose accuracy in specific scenarios while preserving the original pose accuracy.





\section{METHOD}

\begin{figure*}[t]
  \centering
  \includegraphics[width=1\linewidth, height=0.5\linewidth]{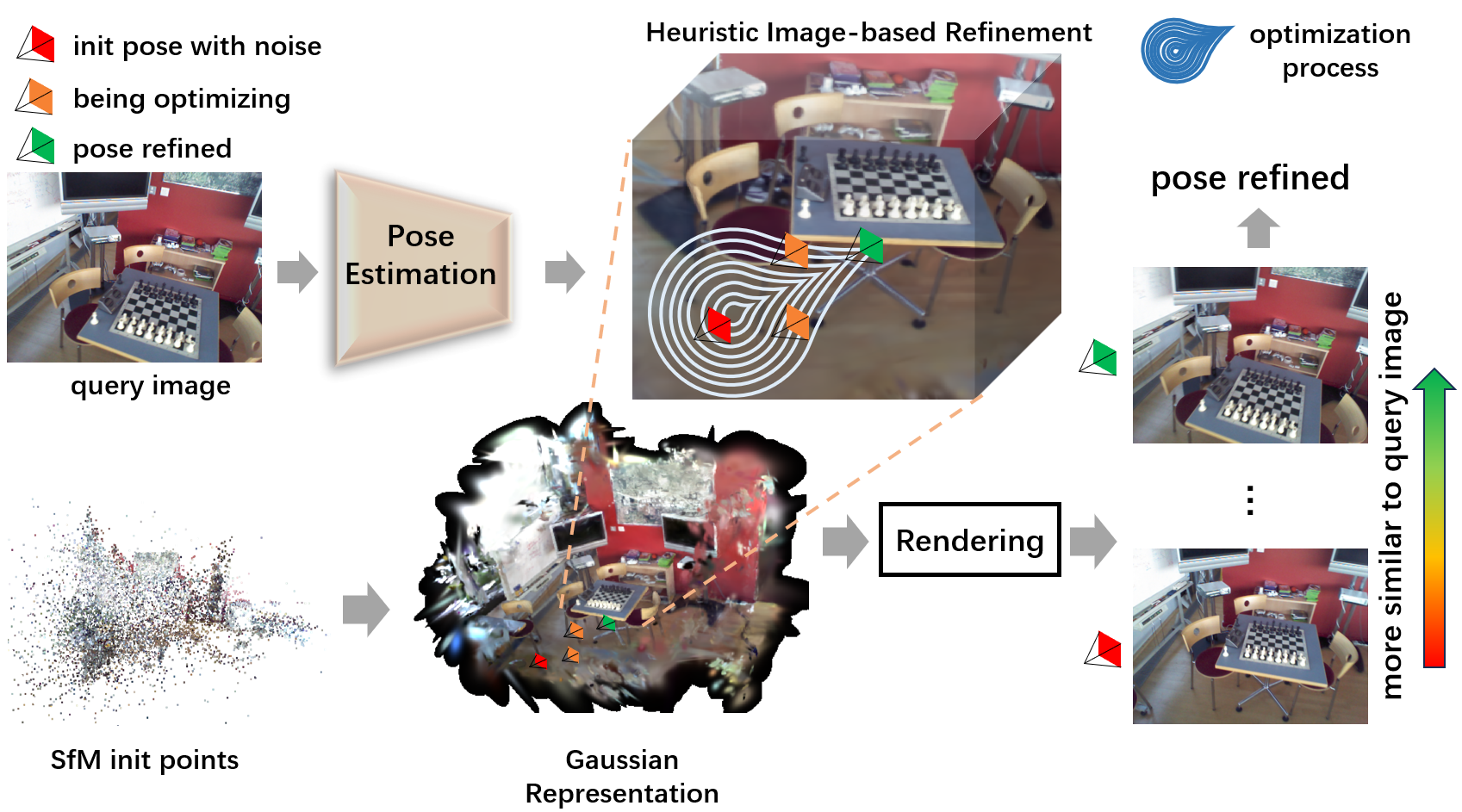}
  \caption{Overview of HGSLoc. Coarse pose estimates are generated by a pre-trained pose estimator, while high-quality reconstructed scenes are obtained through Gaussian densification. The rendered image of the coarse pose in the scene differs significantly from the query image. After applying the heuristic optimization algorithm, the rendered image aligns much more closely with the query image, resulting in a more accurate pose estimate.}
  \label{fig:graph2}
\end{figure*}

In this section, we outline the fundamental principles of the 3D Gaussian Splatting (3DGS) and heuristic refinement strategy, along with their integrated implementation. An overview of our framework is depicted in \cref{fig:graph2}.

\subsection{Explicit Geometric Map} 
3D Gaussian Splatting (3DGS)\cite{kerbl20233d} is a method for representing and rendering three-dimensional scenes. It models the distribution of objects within a scene using 3D Gaussian functions and approximates object surface colors through spherical harmonic coefficients. This method not only delivers an accurate depiction of scene geometry but also effectively captures and renders the lighting and color variations. In 3DGS, each primitive is characterized by a three-dimensional covariance matrix $\boldsymbol{\Sigma}_i\in\mathbb{R}^{3\times3}$ and mean value $\mathbf{\mu}_i\in\mathbb{R}^3$:
$$
g_i(\mathbf{x})=e^{-\frac12(\mathbf{x}-\mathbf{\mu}_i)^\top\boldsymbol{\Sigma}_i^{-1}(\mathbf{x}-\mathbf{\mu}_i)} \eqno{(1)}
$$
where $\boldsymbol{\Sigma}=\mathbf{RSS}^\top\mathbf{R}^\top $, $\mathbf{R}\in\mathbb{R}^{3\times3}$ represents the rotation, $\mathbf{S}\in\mathbb{R}^3$ represents the anisotropy scale.

When projecting onto the viewing plane, 3D Gaussian Splatting (3DGS) utilizes a 2D Gaussian directly, rather than performing the axial integral of a 3D Gaussian. This approach addresses the computational challenge of requiring a large number of samples by limiting the computation to the number of Gaussians, thereby enhancing efficiency. The projected 2D covariance matrix and means are $\boldsymbol{\Sigma}^{\prime}=\mathbf{J}\mathbf{W}\boldsymbol{\Sigma}\mathbf{W}^\mathbf{T}\mathbf{J}^\mathbf{T}$ and $\mathbf{\mu}^{\prime}=\mathbf{JW\mu}$, respectively, where $\textbf{W}$ represents the transformation from the world coordinate system to the camera coordinate system and $\textbf{J}$ denotes the radial approximation of the Jacobian matrix for the projection transformation.

During the rendering phase, spatial depth and tile ID are utilized as key values to sort the Gaussian primitives using GPU-based ordering. Subsequently, the color of each pixel is computed based on the volume rendering formula:
$$
C=\sum_{i\in\mathcal{N}}\mathbf{c}_ip_i\alpha_i\prod_{j=1}^{i-1}(1-p_i\alpha_j) \eqno{(2)}
$$
Where:
$$
p_i=e^{-\frac12(x-\mu_i^{\prime})^T\Sigma^{\prime-1}(x-\mu_i^{\prime})} \eqno{(3)}
$$
$$
\alpha_{2d}=1-\exp\left(-\frac{\alpha_{3d}}{\sqrt{\det(\Sigma_{3d})}}\right) \eqno{(4)}
$$

A major advantage of 3D Gaussian Splatting (3DGS) is its efficient rendering speed. By leveraging CUDA kernel functions for pixel-level parallel processing, 3DGS achieves rapid training and rendering. Additionally, 3DGS employs adaptive control strategies to accommodate objects of various shapes, enhancing both the accuracy and efficiency of rendering. This results in high-quality reconstructed scenes and more realistic new-view images, which provide opportunities for further advancements in pose accuracy.

\subsection{Heuristic Algorithm Implementation}



Heuristic approaches\cite{bulitko2007graph} are often implemented to path planning and graph search, combining the strengths of depth-first search (DFS) and breadth-first search (BFS). It has been widely applied to various real-world problems, including game development, robotic navigation, and geographic information systems (GIS). The primary goal of the heuristic algorithm is to efficiently find the optimal path from an initial node to a goal node, where each node represents a state within the search space. The algorithm relies on an evaluation function, $f(n)$, to prioritize nodes for expansion. This function typically consists of two components:
$$
f(n)=g(n)+h(n) \eqno{(5)}
$$
Where $g(n)$ function is the actual cost from the start node to the current node; $h(n)$ function is the estimated cost from the current node to the target node.

The core idea of the heuristic algorithm is to minimize the number of expanded nodes by guiding the search direction using a heuristic function, $h(n)$, while ensuring the least costly path. The heuristic function must satisfy two important properties: Admissibility and Consistency. Admissibility ensures that $h(n)$ never overestimates the cost of traveling from node $n$ to the target node. Consistency requires that for any node $n$ and its neighboring node $n'$, the heuristic function satisfies the following condition:
$$
h(n)\leq g(n,n^\prime)+h(n^\prime) \eqno{(6)}
$$
Where $g(n,n')$ denotes the actual cost from $n$ to $n'$, which ensures that the algorithm does not repeatedly return to an already expanded node. The algorithm has Optimality and Completeness, i.e., it is guaranteed to find the most optimal path from the start node to the goal node, and for a finite search space, the algorithm always finds a solution.

We use 3DGS as a new-viewpoint image renderer with the goal of finding a more suitable pose within a certain range around the initial pose. A pose is characterized by $(q_w,q_x,q_y,q_z,t_x,t_y,t_z)$, where $q_i$ represent quaternion of a rotation and $t_i$ represent translation. We set the rotation and translation variations ${\delta_q}_i$ and ${\delta_t}_i$, and the current node is transformed to other neighboring nodes by different variations. The pose can be viewed as nodes in the search space, while the transitions between different pose correspond to edges in the graph, and this process can be viewed as expanding nodes in the search graph. In this application, the key to the heuristic algorithm is to design a reasonable cost function. We design the actual cost of a child node as the sum of the actual cost of the current node and the length of the path to the child node, and the estimated cost as the difference value between the rendered image and the query image corresponding to the pose of the current node:

$$
g(n_{child})=g(n_{current})+1 \eqno{(7)}
$$
$$
h(n_{child})=\Sigma|I_q-{I_n}_{child}| \eqno{(8)}
$$
Where $I_q$ represents the current query image and ${I_n}_{child}$ represents the rendering image of current child node. 

The heuristic function effectively guides the algorithm toward the optimal pose, ultimately identifying the pose that produces a rendered image most similar to the query image. We provide the pseudo-code for the algorithm's implementation in \cref{tab:A-star_algorithm}. In this pseudo-code, OpenList is used to store nodes awaiting expansion, while CloseList contains nodes that have already been expanded.

\begin{table}[h]
\caption{Heuristic pose optimization strategy}
\label{tab:A-star_algorithm}
\begin{center}
\begin{tabular}{ll}
\hline
Heuristic Algorithm\\
\hline
while openList is not empty:\\
\quad\quad 1. pop top node with $min(f(n))$ from openList.\\
\quad\quad 2. if top is destination node: \\
\quad\quad\quad\quad break \\
\quad\quad 3. closeList.push(top) \\
\quad\quad 4. for each child node of top: \\
\quad\quad\quad\quad if child in closeList: \\
\quad\quad\quad\quad\quad\quad continue \\
\quad\quad\quad\quad computes the $cost_{tentative}$ from the start node to child. \\
\quad\quad\quad\quad if child not in openList: \\
\quad\quad\quad\quad\quad\quad $g(n_{child})=g(n_{current})+1$ \\
\quad\quad\quad\quad\quad\quad $h(n_{child})=\Sigma|I_q-{I_n}_{child}|$ \\
\quad\quad\quad\quad\quad\quad openList.push(child) \\
\quad\quad\quad\quad elif $cost_{tentative} < g(n_{child})$ : \\
\quad\quad\quad\quad\quad\quad $g(n_{child})$ = tentative cost \\
\quad\quad\quad\quad\quad\quad heap adjustments \\

\hline
\end{tabular}
\end{center}
\end{table}

\section{Experiment}

In this section, we compare and analyze the coarse pose with the optimized pose, including pose accuracy and precision.

\begin{table*}[t]
\scriptsize
\renewcommand\arraystretch{1.1}
\setlength{\tabcolsep}{11.5pt}
\caption{We present the results of comparison experiments on the 7Scenes dataset, highlighting the median translation and rotation errors (cm/°) of the pose relative to the ground truth (GT) pose for various methods across seven scenes. The best results are indicated in bold. "NRP" refers to Neural Render Pose Estimation.}
\label{tab:7scenes_pose_error}
\begin{center}
\begin{tabular}{c|c|ccccccc|c}
\hline
 & Method & chess & fire & heads & office & pumpkin & redkitchen & stairs & Avg.↓[cm/°]\\
\hline
APR & Marepo\cite{chen2024map} & 1.9/0.83 & 2.3/0.91 & 2.2/1.27 & 2.8/0.93 & 2.5/0.88 & 3.0/0.99 & 5.8/1.50 & 2.9/1.04 \\
\hline
SCR & ACE\cite{brachmann2023accelerated} & 0.6/0.18 & 0.8/0.31 & 0.6/0.33 & 1.1/0.28 & 1.2/0.22 & 0.8/\textbf{0.20} & 2.9/0.81 & 1.1/0.33 \\
\hline
\multirow{4}{*}{NRP} & HR-APR\cite{liu2024hr} & 2.0/0.55 & 2.0/0.75 & 2.0/1.45 & 2.0/0.64 & 2.0/0.62 & 2.0/0.67 & 5.0/1.30 & 2.4/0.85 \\
 & NeRFMatch\cite{zhou2024nerfect} & 0.9/0.3 & 1.1/0.4 & 1.5/1.0 & 3.0/0.8 & 2.2/0.6 & 1.0/0.3 & 10.1/0.7 & 2.8/0.73 \\
 & Marepo+HGSLoc & 0.7/0.33 & 1.4/0.62 & 1.5/0.92 & 2.2/0.70 & 1.8/0.46 & 2.2/0.63 & 4.8/1.34 & 2.1/0.71 \\
 & ACE+HGSLoc & \textbf{0.5}/\textbf{0.17} & \textbf{0.6}/\textbf{0.25} & \textbf{0.5}/\textbf{0.29} & \textbf{1.0}/\textbf{0.25} & \textbf{1.1}/\textbf{0.21} & \textbf{0.7}/\textbf{0.20} & \textbf{2.8}/\textbf{0.69} & \textbf{1.0}/\textbf{0.29} \\

\hline
\end{tabular}
\end{center}
\end{table*}

\subsection{Implementation}

The deep learning framework employed in this work is PyTorch\cite{paszke2019pytorch}. Each scene is reconstructed using 3D Gaussian Splatting (3DGS) with 30,000 training iterations, running on RTX 4090 GPUs. For the 7Scenes dataset, we adopt the SfM ground truth (GT) provided by \cite{brachmann2021limits}. 

\subsection{Datasets, Metrics and Baselines}

\paragraph{Datasets}
We evaluated our method on two public datasets: 7Scenes and Deep Blending. In the case of the 7Scenes datasets\cite{glocker2013real,shotton2013scene}, the official test lists were used as query images, while the remaining images were utilized for training. For the Deep Blending dataset\cite{hedman2018deep}, we selected four scenes and constructed a test image set following the 1-out-of-8 approach suggested by Mip-NeRF\cite{barron2021mip}.

\paragraph{Evaluation Metrics} 
We show the median rotation and translation errors, and also provide the ratio of pose error within 1cm/1°.

\paragraph{Benchmark}
Our approach is built upon an initial coarse pose estimation. For the APR\cite{kendall2015posenet,kendall2016modelling,kendall2017geometric,wang2020atloc,chen2021direct,chen2022dfnet,shavit2021learning,chen2024map} framework, we have selected the widely recognized Marepo\cite{chen2024map} method as the benchmark for comparison. Similarly, for the SCR\cite{brachmann2021visual,brachmann2023accelerated,wang2024glace} framework, we have chosen the classical ACE\cite{brachmann2023accelerated} method as the benchmark for comparison.

\subsection{Analysis of results}

\paragraph{7Scenes dataset}
For the 7Scenes dataset, we evaluate the performance of Marepo\cite{chen2024map} and ACE\cite{brachmann2023accelerated} after incorporating HGSLoc. \cref{tab:7scenes_pose_error} demonstrates that our method effectively reduces the error in the coarse pose estimates obtained from both Marepo and ACE. Compared to other NRP methods, our approach achieves results with smaller relative pose errors. Furthermore, \cref{tab:7scenes_pose_ratio} presents the ratio of query images with relative pose errors of up to 1 cm and 1°, showing significant improvements after applying the HGSLoc framework. This indicates that our method efficiently optimizes cases involving small relative pose errors, further enhancing accuracy.

\begin{table}[h]
\renewcommand\arraystretch{1.1}
\setlength{\tabcolsep}{21pt}
\scriptsize
\centering
\caption{We present the average percentage of pose errors within 1 cm and 1° on the 7Scenes dataset. "NRP" denotes neural render pose estimation.}
\label{tab:7scenes_pose_ratio}
\begin{center}
\begin{tabular}{c|c|c}
\hline
 & Methods & Avg.↑[1cm,1°] \\
\hline
APR & Marepo\cite{chen2024map} & 6.2 \\
SCR & ACE\cite{brachmann2023accelerated} & 53.7 \\
NRP & Marepo+HGSLoc & 25.8 \\
NRP & ACE+HGSLoc & 59.2 \\
\hline
\end{tabular}
\end{center}
\end{table}

\paragraph{Deep Blending dataset} 
We selected two indoor scenes, "Playroom" and "DrJohnson", and two outdoor scenes, "Boats" and "NightSnow", for testing. For both the Marepo\cite{chen2024map} and ACE\cite{brachmann2023accelerated} methods, we observed that the coarse pose errors were significantly large. This may be attributed to the higher complexity of the Deep Blending dataset compared to the 7Scenes datasets, as well as the limited training data, which may have prevented model convergence. Consequently, we utilized an alternative method (HLoc\cite{sarlin2019coarse}) that leverages point clouds to obtain an initial pose estimate and compared the results. As shown in \cref{tab:db_pose_hloc}, the improvement from boosting is not pronounced, likely due to the high image quality of the Deep Blending dataset, which already provided relatively accurate preliminary poses with the HLoc framework. To better demonstrate the effectiveness of our pose optimization method, \cref{tab:db_noisy_pose} introduces various levels of step noise, making the visualization results more intuitive.

\begin{table}[h]
\renewcommand\arraystretch{1.1}
\setlength{\tabcolsep}{24pt}
\scriptsize
\centering
\caption{ We present the median translation and rotation errors (cm/°) for both the initial estimated pose and the optimized pose relative to the GT pose.}
\label{tab:db_pose_hloc}
\begin{center}
\begin{tabular}{c|c|c}
\hline
 & init error & refine error \\
\hline
Playroom & 0.7/0.060 & 0.6/0.059 \\
DrJohnson & 0.3/0.055 & 0.3/0.054 \\
Boats & 0.5/0.016 & 0.4/0.013\\
NightSnow & 0.2/0.024 & 0.2/0.019\\
\hline
\end{tabular}
\end{center}
\end{table}

\begin{table}[h]
\renewcommand\arraystretch{1.1}
\setlength{\tabcolsep}{13pt}
\scriptsize
\centering

\caption{We show the median translation and rotation errors (m/°) for the poses with noise and for the poses after optimization. (q2, t1) denotes the introduction of noise at the percentile of qvec, decile of tvec, and the rest is the same.}
\label{tab:db_noisy_pose}
 (a) Playroom
\begin{center}
\begin{tabular}{c|c|c|c|c}
\hline
 & noise error & refine error & tvec↑ & qvec↑\\
\hline
q2, t1 & 0.81/7.79 & 0.29/2.72 & 64.2\% & 65.1\% \\
q2, t2 & 0.31/8.42 & 0.16/1.81 & 48.4\% & 78.5\% \\
q3, t3 & 0.03/0.81 & 0.02/0.26 & 33.3\% & 67.9\% \\
\hline
\end{tabular}
\end{center}

 (b) DrJohnson
\begin{center}
\begin{tabular}{c|c|c|c|c}
\hline
 & noise error & refine error & tvec↑ & qvec↑\\
\hline
q2, t1 & 0.68/7.81 & 0.15/1.87 & 77.9\% & 76.1\% \\
q2, t2 & 0.33/7.86 & 0.13/2.21 & 60.6\% & 71.9\% \\
q3, t3 & 0.03/0.72 & 0.01/0.21 & 66.7\% & 70.8\% \\
\hline
\end{tabular}
\end{center}

 (c) Boats
\begin{center}
\begin{tabular}{c|c|c|c|c}
\hline
 & noise error & refine error & tvec↑ & qvec↑\\
\hline
q2, t1 & 0.62/6.41 & 0.01/0.02 & 98.4\% & 99.7\% \\
q2, t2 & 0.26/9.17 & 0.17/2.46 & 34.6\% & 73.2\% \\
q3, t3 & 0.04/0.71 & 0.01/0.15 & 75.0\% & 78.9\% \\
\hline
\end{tabular}
\end{center}

 (d) NightSnow
\begin{center}
\begin{tabular}{c|c|c|c|c}
\hline
 & noise error & refine error & tvec↑ & qvec↑\\
\hline
q2, t1 & 0.49/7.79 & 0.01/0.03 & 98.0\% & 99.6\% \\
q2, t2 & 0.29/7.62 & 0.07/0.58 & 75.9\% & 92.4\% \\
q3, t3 & 0.03/0.69 & 0.01/0.10 & 66.7\% & 85.5\% \\
\hline
\end{tabular}
\end{center}

\end{table}

As shown in \cref{tab:joint_H}, to further demonstrate the effectiveness of our method, we compare it with an alternative joint optimization strategy\cite{fan2024instantsplat}. For this comparison, a noise level of $1\times10^{-3}$ granularity is introduced to the initial pose. Our method employs heuristic optimization based on high-quality scene reconstruction obtained through the 3DGS\cite{kerbl20233d} method, whereas the alternative strategy jointly optimizes both the scene reconstruction and the initial noisy
pose\cite{fan2024instantsplat}.

\begin{table}[h]
\renewcommand\arraystretch{1.1}
\setlength{\tabcolsep}{15pt}
\scriptsize
\centering
\caption{ We show the median translation and rotation errors (m/°) for heuristic optimization and joint optimization strategies.}
\label{tab:joint_H}
\begin{center}
\begin{tabular}{c|c|c|c}
\hline
 & init error & joint error & heuristic error \\
\hline
Playroom & 0.03/0.81 & 0.02/0.42 & 0.02/0.26 \\
DrJohnson & 0.03/0.72 & 0.02/0.47 & 0.01/0.21 \\
Boats & 0.04/0.71 & 0.01/0.19 & 0.01/0.15 \\
NightSnow & 0.03/0.69 & 0.02/0.36 & 0.01/0.10 \\
\hline
\end{tabular}
\end{center}
\end{table}

\begin{figure}[t]
  \centering
  \includegraphics[width=\linewidth]{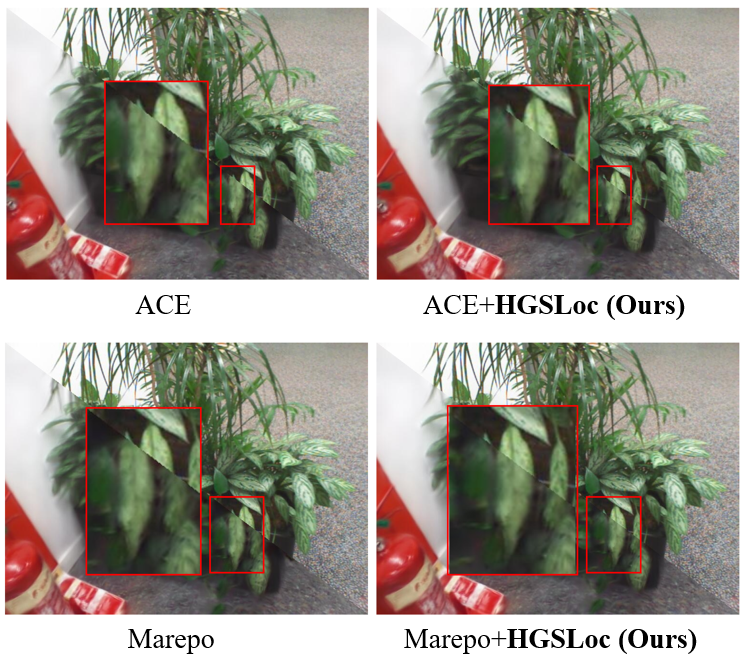}
  \caption{HGSLoc demonstrates a significant optimization effect on the coarse poses obtained using the ACE and Marepo methods. Each subimage is divided by a diagonal line: the rendered image from the pose is shown in the bottom left part, while the GT image is shown in the top right part. The rendered images corresponding to the ACE and Marepo methods exhibit substantial misalignment with the GT images. To facilitate a clearer comparison, we provide a zoomed-in view of the image, highlighted within the red box.}
  \label{fig:7scenes_compare}
\end{figure}

\begin{figure}[t]
  \centering
  \includegraphics[width=\linewidth]{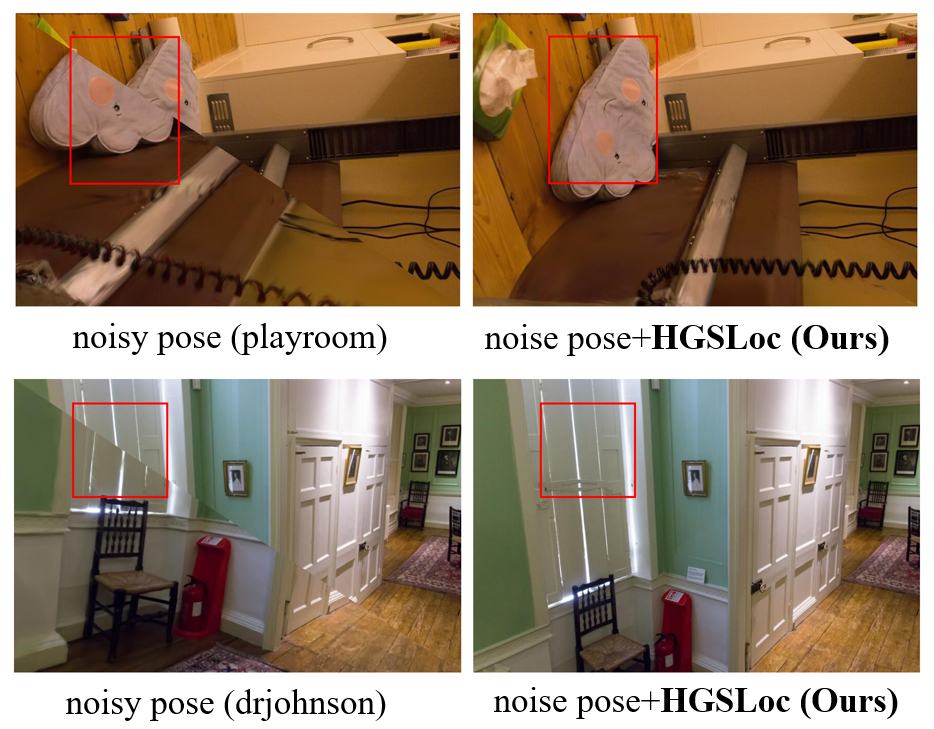}
  (a) Indoor Scenes
  \includegraphics[width=\linewidth]{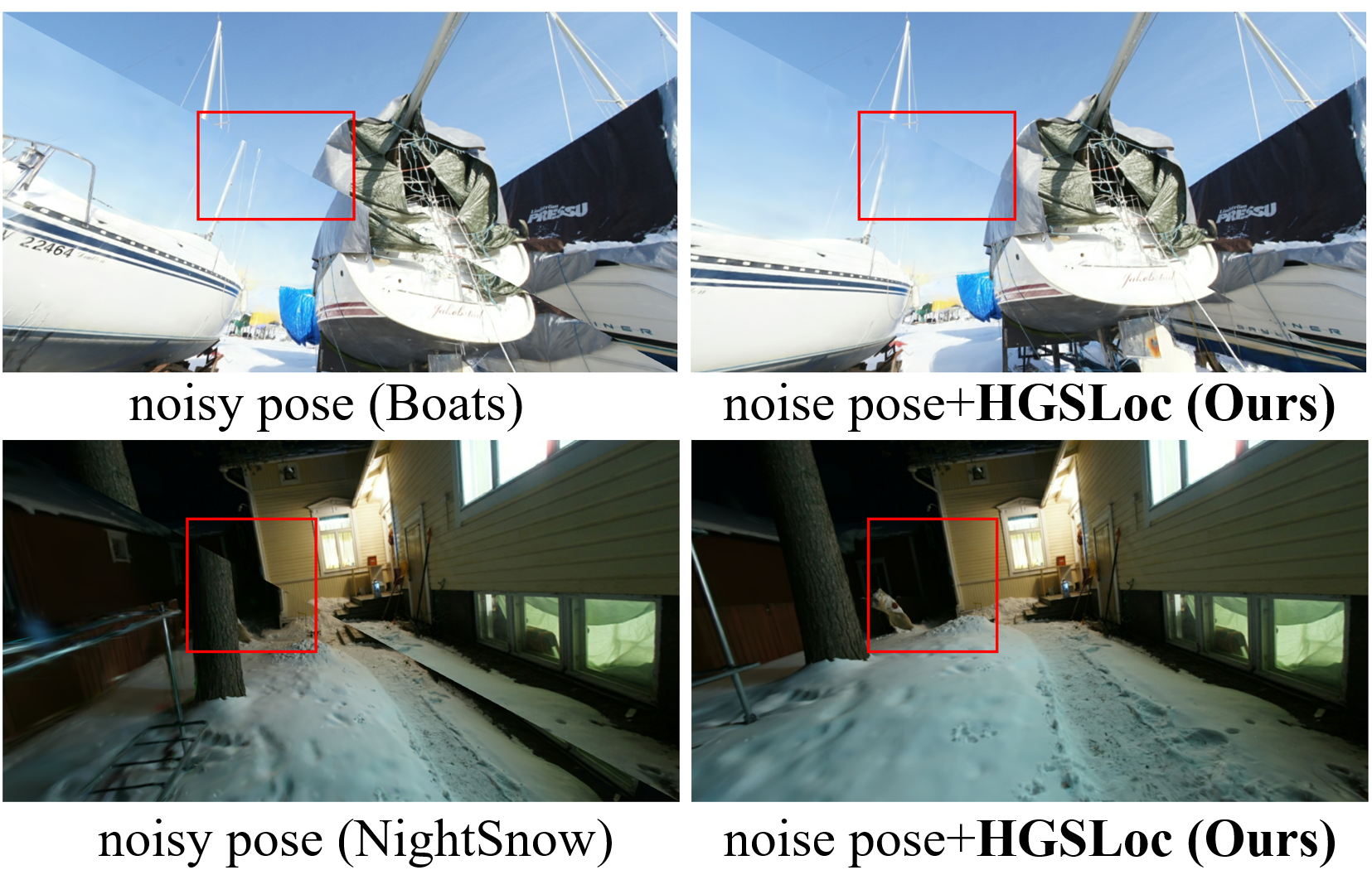}
  (b) Outdoor Scenes
  \caption{Each subimage is divided by a diagonal line, with the image rendered by the estimated pose on the lower left and the GT image on the upper right. The diagonal lines in the optimized comparison image appear less distinct, reflecting improved alignment with the GT image. HGSLoc demonstrates its effectiveness in refining pose estimation, achieving precise values while mitigating the impact of noise.}
  \label{fig:db_compare}
\end{figure}

\paragraph{Qualitative Analysis}
By inputting the pose into the 3D reconstructed scene, we generate a rendered image that visualizes the pose. Each query image corresponds to the GT pose, and the discrepancy between the estimated pose and the GT pose is reflected in the rendered images from various viewpoints. To better analyze errors and optimization improvements, we select viewpoints where significant accuracy gains are observed. \cref{fig:7scenes_compare} demonstrate that, when using our framework on the 7Scenes datasets, the rendered images more closely match the GT images. \cref{fig:db_compare} illustrates the results of applying our framework to noisy poses in the Deep Blending dataset, showing that our method effectively refines the original pose, resulting in rendered images that closely resemble the GT images. However, experiments on the Cambridge Landmarks dataset\cite{kendall2015posenet} yielded suboptimal results, likely due to its limited viewpoint coverage, which leads to gaps in the 3D reconstruction of the scenes. In contrast, the 360° closed-loop image structure of the Deep Blending dataset enhance the effectiveness of pose optimization.

\paragraph{Ablation study}
In our method, we use the sum of pixel-by-pixel differences as the heuristic function. To demonstrate the effectiveness of this heuristic function, \cref{tab:different_H} compares the results obtained using Peak Signal-to-Noise Ratio (PSNR) and Structural Similarity Index (SSIM) as alternative heuristic functions. Higher values of PSNR and SSIM indicate better image quality and structural similarity, whereas we would like to see them take the opposite number as the value of the heuristic function is as small as possible. To illustrate the impact of different heuristic functions more clearly, we applied these comparisons to the Deep Blending dataset, which introduces a relatively large level of noise.

$$
h_1(n_{child})=100-PSNR(I_q,{I_n}_{child}) \eqno{(9)}
$$
$$
h_2(n_{child})=1.0-SSIM(I_q,{I_n}_{child}) \eqno{(10)}
$$

\begin{table}[h]
\caption{We show the median translation and rotation errors (m/°) for poses with noise and for poses after optimization using different heuristic functions.}
\label{tab:different_H}
\begin{center}
\begin{tabular}{c|c|c|c|c}
\hline
 & noise error & H(Sum of Diff) & H(PSNR) & H(SSIM) \\
\hline
Playroom & 0.81/7.79 & 0.29/2.72 & 0.76/6.29 & 0.87/6.83 \\
DrJohnson & 0.68/7.81 & 0.15/1.87 & 0.60/6.61 & 0.65/7.59 \\ 
Boats & 0.62/6.41 & 0.01/0.02 & 0.44/3.93 & 0.59/5.97 \\
NightSnow & 0.49/7.79 & 0.01/0.03 & 0.49/7.03 & 0.46/7.25 \\
\hline
\end{tabular}
\end{center}
\end{table}

\section{LIMITATIONS AND FUTURE WORK}

While our method successfully achieves the desired results, it still has certain limitations. During the optimization process, typically 300–400 nodes need to be expanded, with multiple Gaussian renderings performed for each expansion, which significantly constrains the method’s real-time performance. Future work will focus on optimizing pose representation to reduce the number of Gaussian renderings required per node expansion, thereby improving computational efficiency and enhancing the method’s applicability to real-time scenarios.

\section{CONCLUSIONS}

This study introduces a lightweight, plug-and-play visual localization framework that integrates heuristic refinement with 3D reconstruction to enhance pose accuracy, achieving state-of-the-art performance on two datasets. Compared to NeRF-based methods\cite{zhou2024nerfect}, it achieves superior localization accuracy by efficiently refining coarse estimations using heuristic functions. The modular approach not only  reduces dependence on complex neural network training, but also demonstrates robust performance in noisy environments. By combining heuristic refinement with a 3D Gaussian distribution, this approach offers a novel and effective solution, providing valuable reference for the future development of visual localization systems.













\balance
\bibliographystyle{IEEEtran} 
\bibliography{root} 

\end{document}